\documentclass[conference]{IEEEtran}
\IEEEoverridecommandlockouts

\usepackage{cite}
\usepackage{amsmath,amssymb,amsfonts}
\usepackage{algorithmic}
\usepackage{graphicx}
\usepackage{textcomp}
\usepackage{xcolor}
\usepackage{makecell}
\usepackage{booktabs} 
\usepackage{amsmath} 
\usepackage{amsfonts} 
\usepackage{algorithm}
\usepackage{graphicx} 
\usepackage{array} 
\usepackage{algorithmic}
\usepackage{subcaption}
\usepackage{graphicx}
\usepackage{textcomp}
\usepackage{xcolor}
\usepackage{rotating} 
\usepackage{multirow}
\usepackage{tabularx}
\usepackage{diagbox}
\usepackage{caption}
\usepackage{footnote}
\usepackage{enumitem}
\usepackage{hyperref}
\def\BibTeX{{\rm B\kern-.05em{\sc i\kern-.025em b}\kern-.08em
    T\kern-.1667em\lower.7ex\hbox{E}\kern-.125emX}}

\begin{document}

\newcommand{\rqin}[1]{\textcolor{black}{#1}}
\newcommand{\dcheng}[1]{\textcolor{red}{#1}}
\newcommand{\jx}[1]
{\textcolor{blue}{jx: #1}}

\title{NVCiM-PT: An NVCiM-assisted Prompt Tuning Framework for Edge LLMs}

\author{Ruiyang Qin$^{1, 2}$, Pengyu Ren$^{1}$, Zheyu Yan$^{1}$, Liu Liu$^{1}$, Dancheng Liu$^{2}$, Amir Nassereldine$^{2}$, 
\\ Jinjun Xiong$^{2}$, Kai Ni$^{1}$, Sharon Hu$^{1}$, Yiyu Shi$^{1}$
\\ $^{1}$University of Notre Dame $^{2}$University at Buffalo--SUNY}



\maketitle

\begin{abstract}
Large Language Models (LLMs) deployed on edge devices, known as edge LLMs, need to continuously fine-tune their model parameters from user-generated data under limited resource constraints. However, most existing learning methods are not applicable for edge LLMs because of their reliance on high resources and low learning capacity.
Prompt tuning (PT) has recently emerged as an effective fine-tuning method for edge LLMs by only modifying a small portion of LLM parameters, but it suffers from user domain shifts, resulting in repetitive
training and losing resource
efficiency.
Conventional techniques to address domain shift issues often involve
complex neural networks and sophisticated training, which are
incompatible for PT for edge LLMs.
Therefore, an open research question is how to address domain shift issues for edge LLMs with limited resources. 
In this paper, we propose a prompt tuning framework for edge LLMs, exploiting the benefits offered by non-volatile computing-in-memory (NVCiM) architectures. 
We introduce a novel NVCiM-assisted PT framework, where we narrow down the core operations to matrix-matrix multiplication, which can then be accelerated by performing in-situ computation on NVCiM. To the best of our knowledge, this is the first work employing NVCiM to improve the edge LLM PT performance.

\end{abstract}


\section{Introduction}
\label{sec:intro}
Large Language Models (LLMs) are primarily deployed on centralized cloud platforms \cite{qin2024fl, qin2024language, bevilacqua2023automated}, raising significant concerns about user privacy and trustworthiness \cite{neel2023privacy}. These issues are particularly critical in domains such as healthcare \cite{Karabacak_Margetis_2023}, AI companionship \cite{xu2023large}, and personal assistance \cite{li2024personal}, where protecting user data and ensuring model reliability are crucial. To address these challenges, there is a shift towards developing personalized LLMs capable of generating tailored responses. These models are designed for deployment on edge devices like Jetson Orin, hence they called \textit{Edge LLMs}. 
By enabling local learning and adaptation, Edge LLMs offer a promising solution to balance powerful AI capabilities with enhanced privacy protection, potentially transforming how we interact with and benefit from LLMs.

To better customize the model for individual users, edge LLMs must learn from their interaction with users. Under the constraint of memory capacity and computational power on edge devices, existing works mainly take two types of approaches to enable LLM learning on edge: retrieval-augmented generation (RAG) \cite{qin2024robust} and low-rank adaption (LoRA) \cite{qin2023enabling}. However, recent studies on edge LLMs \cite{qin2024empirical} found that RAG offers limited improvement due to the small size of pre-trained parameters on edge LLM, while LoRA's high resource consumption can prevent it from learning a broad range of user-generated data. 

Prompt tuning \cite{li2021prefix} (PT), as an alternative with higher learning capacity than RAG and lower resource usage than LoRA, can be a good learning method for edge LLM. 
\rqin{When the PT is applied to an LLM, a set of virtual tokens will be produced. The PT typically adjusts less than 0.01\% of the model's parameters. By concatenating the set of virtual tokens with user  inputs' embedding, PT can better align user inputs with pre-trained LLMs.} 
However, PT's application to edge LLMs faces significant challenges. Despite the smaller number of trainable parameters, PT substantially increases training and inference time due to the transformer's quadratic complexity \cite{shi2023dept}. 
Furthermore, edge devices' limited data buffer sizes exacerbate domain shift issues as users switch between tasks. This is in direct contrast to cloud-based PT that can create versatile ``one-for-all'' (\textbf{one4all}) \cite{li2021prefix} set of virtual tokens on a large dataset.



\begin{figure}[t!]
  \centering
  \begin{subfigure}[b]{0.24\textwidth}
    \centering
    \includegraphics[width=\textwidth]{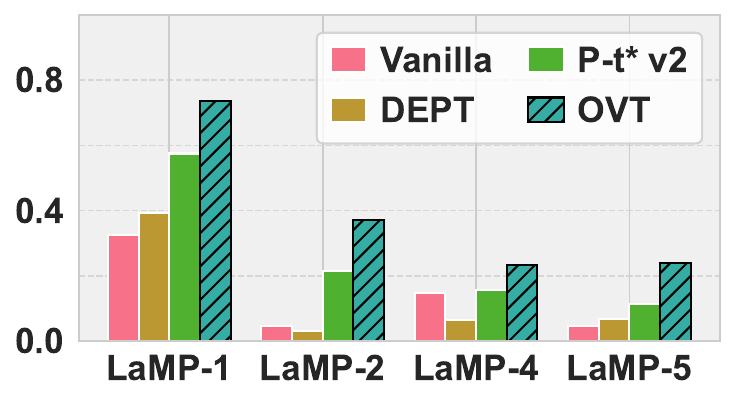}
    \captionsetup{font=small}  
    \vspace{-4ex}
    \caption{Gemma-2B}
    \label{fig:prelim_gemma_2b}
  \end{subfigure}
  \begin{subfigure}[b]{0.24\textwidth}
    \centering
    \includegraphics[width=\textwidth]{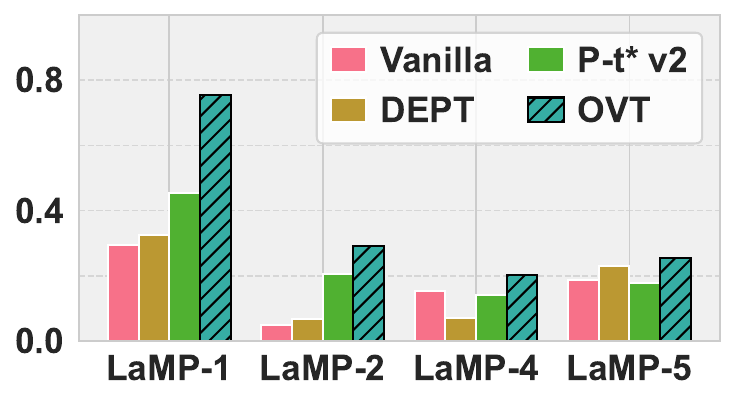}
    \captionsetup{font=small}  
    \vspace{-4ex}
    \caption{Phi-2}
    \label{fig:prelim_phi_2}
  \end{subfigure}
  \captionsetup{font=small}  
  \vspace{-3ex}
  \caption{Edge LLM performance comparison 
  for two LLMs (Gemma-2B and Phi-2)
  across four datasets on 
  four prompt tuning methods---\textbf{Vanilla}, \textbf{DEPT},  \textbf{P-tuning v2} (P-t* v2), and 
  prefix tuning with \textbf{OVT} (optimal sets of virtual tokens).
  }
  \vspace{-3ex}

  \label{tab:one4all}
\end{figure}


Previous research \cite{nassereldine2024pi} suggests that each data sample can be associated with an \textbf{O}ptimal set of \textbf{V}irtual \textbf{T}okens (\textbf{OVT}). This OVT, derived through PT on that specific sample, can significantly enhance the LLM's performance.
Our comparative analysis of one4all and OVTs across three edge LLMs and five datasets (detailed in Section~\ref{sec:exp}) reveals a striking performance gap. 
As shown in Fig.~\ref{tab:one4all}, we compare \rqin{a modified version of the prefix tuning method \cite{li2021prefix}, which uses OVTs instead of one4all virtual tokens}, with three alternative PT methods---Vanilla \cite{lester2021power}, DEPT \cite{shi2023dept}, and P-tuning v2 \cite{liu2021p}, where the LLM use one4all prompts to deal with all data samples. 
It clearly shows that the prefix PT with OVTs can significantly outperform the three other PT methods. This finding highlights the potential of using OVTs in PT to address the performance downgrade issues due to domain shift, offering a promising direction for edge LLM optimization.

\begin{figure}[t!]
  \centering
  \begin{subfigure}[b]{0.24\textwidth}
    \centering
    \includegraphics[width=\textwidth]{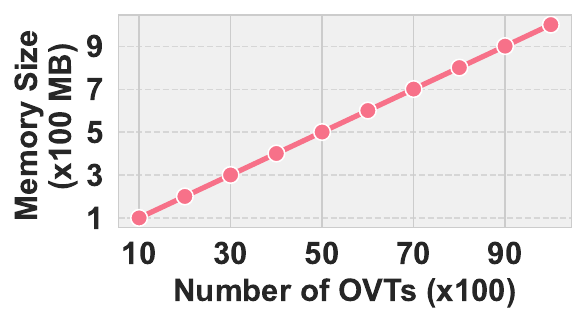}
    \captionsetup{font=small}
    \vspace{-4ex}
    \caption{Increasing memory usage}
    \label{fig:virtual_token_size}
  \end{subfigure}
  \begin{subfigure}[b]{0.24\textwidth}
    \centering
    \includegraphics[width=\textwidth]{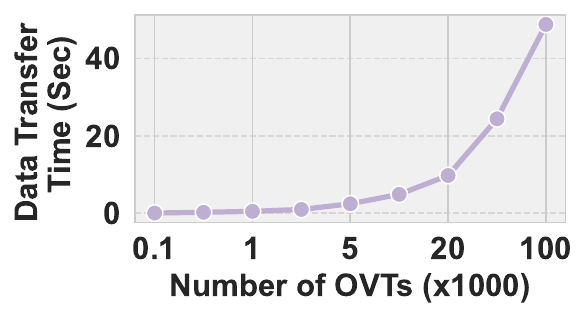}
    \captionsetup{font=small}
    \vspace{-4ex}
    \caption{Increasing data moving time}
    \label{fig:soft_prompt_time}
  \end{subfigure}
  \captionsetup{font=small}  
  \caption{Resource by the storing virtual tokens and data moving.}
\vspace{-4ex}

  \label{fig:combined_analysis}
\end{figure}

Unfortunately, directly applying OVT to address domain shift issues would result in
an inevitable growth of OVT volumes
with the increasing of user-generated
data, which strains the limited
edge resources as shown in Fig.~\ref{fig:virtual_token_size}. 
Furthermore, retrieving the appropriate OVTs can involve either high Dynamic Random Access Memory (DRAM) usage or high latency if data are transferred between solid-state drive (SSD) and DRAM, as shown in Fig.~\ref{fig:soft_prompt_time}.
To address these issues, we notice that Computing-in-Memory (CiM) backed by Non-volatile Memory (NVM) 
i.e., NVCiM as an emerging computing architecture, can integrate computation within memory arrays to reduce data movement and enhance energy efficiency. Hence, NVCiM can be a potential solution to address the two challenges of using the OVT, where each OVT can be stored in NVM while the data input, as a query, can find its corresponding OVT via matrix-matrix multiplication as implemented on the CiM architecture. 

However, there are three main challenges to effectively utilizing NVCiM to assist the PT in using OVT. First, obtaining the virtual tokens for every data sample can be resource-inefficient since the data samples within a certain amount of time can be highly correlated in their domains and user preferences. It remains unknown how to decide when to run prompt tuning and obtain the virtual tokens. Second, NVM device variations alter the stored virtual tokens and impact the PT-based LLM performance. Device variations can be different for different NVMs. 
Third, it is not clear whether the established retrieved method \cite{yin2024ferroelectric} like Max Inner Product Search (MIPS) can still be effective in retrieving the OVT for a given data input.

We propose a framework that enables NVCiM to assist prompt tuning (\textbf{NVCiM-PT}) with OVTs to improve the performance of edge LLMs. Our framework consists of three components: (i) we develop a resource-efficient data selection method to refine and reduce the data samples in the data buffer for prompt tuning; (ii) a noise-aware training method for PT so the obtained virtual tokens can be resilient to NVM device variation; (iii) we design a new retrieval algorithm that can ensure the appropriate virtual tokens to be retrieved in various NVCiMs.

The main contributions of this work are summarized in the following:
\begin{itemize}[leftmargin=*]
    \item We propose the first NVCiM-assisted framework for edge LLMs, unleashing the potential to utilize OVTs in PT. 
    \item We introduce a heuristic search algorithm for retrieving the OVTs with an NVCiM architecture and demonstrate its superior performance.
    \item Our experiments on various datasets show that our proposed framework can improve the edge LLM performance on multiple NVCiM devices by up to 36.7\%, along with up to 120$\times$ improvement of latency and up to 60$\times$ improvement of energy compared to using Jetson Orin CPU.
\end{itemize}


\section{Background}

\subsection{NVMs and their device variations}
\label{sec:NVM}

NVM has been increasingly utilized in CiM architectures for its enhanced energy efficiency \cite{jung2022crossbar}.
Two common NVM types used in CiM are Resistive Random-Access Memory (RRAM) and Ferroelectric Field-Effect Transistors (FeFET). RRAM stores data by changing the resistance across a dielectric material \cite{wan2022compute}, while FeFET utilizes ferroelectric materials to maintain data through polarization states \cite{khan2020future}. Both RRAM and FeFET offer high density, CMOS compatibility, and low read energy; however, they differ in write speeds, voltage requirements, and endurance.


\begin{figure*}[t!]
  \centering
  \includegraphics[trim=0 400 287 0, clip, width=1.\linewidth]{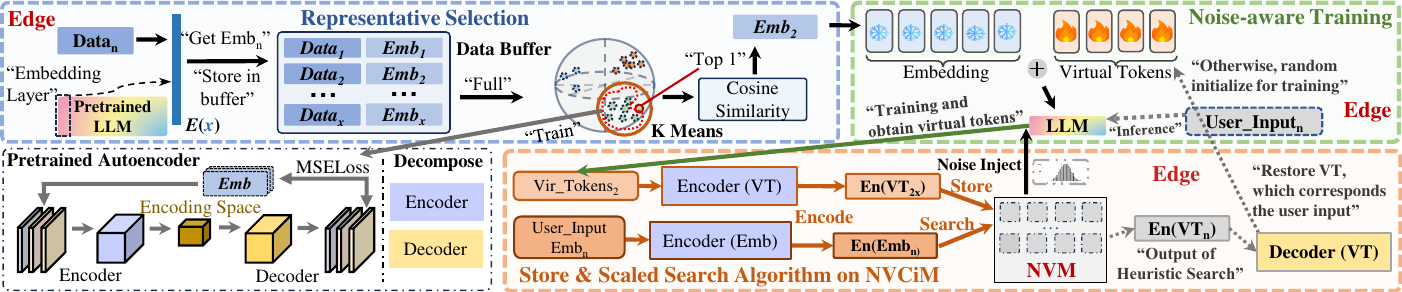}
  \captionsetup{font=small}  
  \caption{Overview of our proposed \textbf{NVCiM}-based \textbf{P}rompt Tuning framework (\textbf{NVCiM-PT}). By co-design, it can utilize different types of NVCiM to improve prompt-tuning-based LLM content generation.}
  \vspace{-3ex}
  \label{fig:method}
\end{figure*}


One drawback of CiM is that it suffers from various sources of variations and noises, two major of which are spatial variations and temporal variations. Spatial variations result from fabrication defects and may have both local and global correlations. FeFET devices also suffer from temporal variations due to the stochasticity in memory switching and aging, which causes fluctuations in conductance when programmed at different times. Measurement results~\cite{yan2022swim, yan2021uncertainty} show that the noise on DNN weights caused by device variations can be modeled as a Gaussian noise with zero mean and each weight associated with a corresponding standard deviation . A detailed representation is given by
\(\mathbf{v} = \mathbf{v_0} + \Delta\mathbf{v}\) with 
\(\Delta\mathbf{v}\sim\mathcal{N}(0,\sigma_v)\),
where $\mathbf{v}$ is the actual weight deployed on the accelerators, $\mathbf{v_0}$ is the target weight value, and $\sigma_v$ is a value measured by the experiments. We collect the measurement results from RRAM and FeFET devices and the specific values will be discussed in Section~\ref{sec:default_setting}.

\subsection{Prompt Tuning}

\rqin{PT improves the performance of LLM by altering the data inputs. Since Fig.~\ref{tab:one4all} has shown that
prefix tuning with OVTs is superior to most state-of-the-art PT methods, we adopt the method of prefix tuning as our PT framework in this work.
Given data inputs, it can produce either a one4all set of virtual tokens or many OVTs with each OVT only corresponding to a specific domain of data inputs. 
In this work, we investigate an NVCiM-backed PT approach. Instead of using a one4all set of virtual tokens, we train and store OVTs in NVM. During inference, we retrieve the appropriate sets of virtual tokens using NVCiM and concatenate them with the input embedding.  
By selecting the most suitable OVT prompt for each input, we can significantly enhance the LLM's performance.}


\section{Proposed Work}
\subsection{Framework Overview}

In this section, we introduce our \textbf{NVCiM}-assisted \textbf{P}rompt \textbf{T}uning framework (\textbf{NVCiM-PT}). This framework modifies the typical PT pipeline from a one-for-all approach to searching for OVTs, designed to effectively leverage NVCiM. NVCiM-PT is specifically tailored for resource-constrained environments where LLMs need to learn on edge devices, enabling the efficient use of OVTs.

When NVCiM-PT is deployed, it operates in two main modes. During the training mode, users accumulate data on their edge device, where the representative selection is involved to produce the training data samples from user-generated data and the LLM runs PT to obtain the OVTs on these data samples. NVCiM-PT incorporates a noise-injection module during PT to enhance noise-resilience of the virtual tokens, which is crucial for subsequent storage on NVM. In the inference mode, where PT is not initiated, users can utilize the NVM-stored virtual tokens for LLM inference. In this mode, NVCiM-PT employs a scaled search module to instantly retrieve the appropriate set of virtual tokens corresponding to each user input. This retrieval process is performed on the NVCiM architecture. The retrieved set of virtual tokens is then concatenated with the user input embedding, and this combination is fed into the LLM for output generation.

NVCiM-PT implements these functionalities by incorporating three components which are detailed in Fig.~\ref{fig:method}. The first component is representative selection (\textbf{RS}) shown in the blue block in Figure~\ref{fig:method}. The representative selection selects the most representative data samples from the data buffer. The second component is noise-aware training (\textbf{NT}) shown in the green block in Figure~\ref{fig:method}. It takes the selected data samples and trains their noise-resilient virtual tokens. The third component is store \& scaled search algorithm (\textbf{SSA}) on NVCiM shown in the orange block in Figure~\ref{fig:method}. It stores the obtained sets of virtual tokens on NVM and searches for the OVT when being given queries. Details of the content in each block are discussed below.

\subsection{Representative Selection (RS)}

Usually, when using prompt tuning for edge LLMs, the ``one4all" set of virtual tokens is obtained from a small range of data samples. In our NVCiM-PT framework, instead of the only one4all set of virtual tokens, multiple OVTs are obtained. While each OVT works well within a small domain of data, the accumulation of these prompts allows our framework to always find the appropriate prompt, given any domain.

We introduce an RS component (blue block in Fig.~\ref{fig:method}) to identify the most representative samples in every small domain of data samples. These data samples are generated during the user-LLM interaction and are temporarily held in a data buffer. \rqin{For the data buffer, it temporarily maintains the user-generated data with annotated user-preference.} When the data buffer is full, RS employs the \( k\text{-means}() \) function (Equation~\ref{eq:kmeans}) to identify the domains for all the stored samples. 

{\footnotesize
\begin{equation}
\label{eq:kmeans}
\{ C_1, C_2, \ldots, C_k \} = k\text{-means}(E, k)
\end{equation}}

\noindent where \( E \) is the set of embeddings of all data samples in the buffer, \( k \) is the number of clusters, and \( C_i \) represents the \( i \)-th cluster (domain). To obtain an appropriate \( k \) value, we employ Equation~\ref{eq:kvalue}. 

{\footnotesize
\begin{equation}
\label{eq:kvalue}
k = \min\left( \max\left( \left\lfloor n_{\min} + s \cdot \log_2\left( \frac{b_s}{b_0} \right) \right\rfloor, n_{\min} \right), n_{\max} \right)
\end{equation}}where \( b_s \) is the buffer size, \( b_0 \) is the base threshold, \( s \) is the scale factor, and \( n_{\min} \) and \( n_{\max} \) are the minimum 
and maximum 
number of clusters, respectively. Each cluster represents one domain. Within each domain, one most representative sample is selected based on the cosine similarity function \( \text{cos\_sim}() \) (Equation~\ref{eq:cosine}). Specifically for \( C_i \), we select the most representative sample \( e_i^* \) as:




{\footnotesize
\begin{equation}
\label{eq:cosine}
e_i^* = \underset{e \in C_i}{\arg\min} \; \text{cos\_sim}(e, \mu(C_i))
\end{equation}}where \( e \) is an embedding in cluster \( C_i \), \( \mu(C_i) \) is the centroid of cluster \( C_i \), \( \text{cos\_sim}(e, \mu(C_i)) \) calculates the cosine similarity between \( e \) and \( \mu(C_i) \).

\subsection{Noise-aware Training (NT)}


From the RS part, we obtain \( k \) representative data samples. For each of these samples, its virtual tokens can be obtained from the Huggingface default prompt tuning method \cite{lester2021power}.
However, storing these sets of virtual tokens in NVM devices introduces vulnerability to device variations, i.e., noise in the values of virtual tokens.
One approach to mitigate the noise impact involves determining the exact noise profile for each individual NVM device and applying a corresponding noise-cancellation mask when writing sets of virtual tokens to the NVM \cite{gong2018signal}. However, this method has significant drawbacks. Accurately measuring the noise profile is both costly and complex due to the non-linear and asymmetric switching behaviors of these devices. Moreover, this approach becomes impractical in the context of user-specific edge-based LLMs, where each NVM device may exhibit a unique noise profile.
Alternatively, obtaining a Gaussian distribution of the noise for the same type of NVM \cite{yan2022reliability} can be easier. Hence, we introduce a noise-aware training method to inject noise based on a Gaussian distribution parameterized by $\sigma$ (sigma) into the virtual tokens during the training process. In this way, noise-aware training (\textbf{NT}) can shape the virtual tokens to be more resilient to certain noise distributions.

Our NT (the green block in Fig.~\ref{fig:method}) begins with the initialization of a random set of virtual tokens \( S \), which is concatenated with the input embedding. The concatenation will undergo the LLM to optimize virtual tokens.
Crucially, we inject simulated noise due to device variation  into the virtual tokens during this process, governed by Equation~\ref{eq:noise_injection}. The noise magnitude depends on the normalized value of each set of virtual tokens element \( \hat{S}_{ij} \).


{\scriptsize
\begin{equation}
\label{eq:noise_injection}\vspace{-3ex}
\begin{gathered}
S' = S + N \cdot \max(|S|), \quad \hat{S} = \frac{S}{\max(|S|)} \\
N_{ij} = \begin{cases} 
\mathcal{N}(0, (\sigma f_1)^2), & \text{if } |\hat{S}_{ij}| > 0.75 \\
\mathcal{N}(0, (\sigma f_2)^2), & \text{if } 0.5 \leq |\hat{S}_{ij}| \leq 0.75 \\
\mathcal{N}(0, (\sigma f_3)^2), & \text{if } 0.25 \leq |\hat{S}_{ij}| < 0.5 \\
\mathcal{N}(0, (\sigma f_4)^2), & \text{if } |\hat{S}_{ij}| < 0.25
\end{cases}
\end{gathered}
\end{equation}}

In this equation, \( S' \) is the noise-injected virtual tokens, \( S \) is the original virtual tokens, \( N \) is the injected noise matrix, and \( \hat{S} \) is the normalized virtual tokens. The noise \( N_{ij} \) follows a Gaussian distribution \( \mathcal{N}(0, (\sigma f)^2) \), where \( \sigma \) is the global noise parameter (\textit{gaussian\_noise\_sigma}), and \( f_1, f_2, f_3, f_4 \) are noise factors corresponding to different intervals of \( |\hat{S}_{ij}| \). The noise is applied to \( S \), scaled by the maximum absolute value of \( S \) to maintain relative magnitude. This noise-aware training approach ensures that the resulting virtual tokens is inherently robust to the device variations expected during NVM storage and retrieval.

\subsection{Store \& Scaled Search Algorithm (SSA) on NVCiM}

We store the OVTs to crossbar arrays after the noise-aware training. 
Meanwhile, we also retrieve the appropriate virtual tokens based on the user input.
To implement these two functionalities we first employ a pre-trained autoencoder to reshape the obtained virtual tokens to fit the NVMs. Then we introduce a scaled search algorithm (\textbf{SSA}) to retrieve the appropriate NVM-stored virtual tokens.


\subsubsection{Store OVTs on NVMs}

\begin{figure}[t!]
  \centering
  \includegraphics[trim=0 375 645 0, clip, width=.95\linewidth]{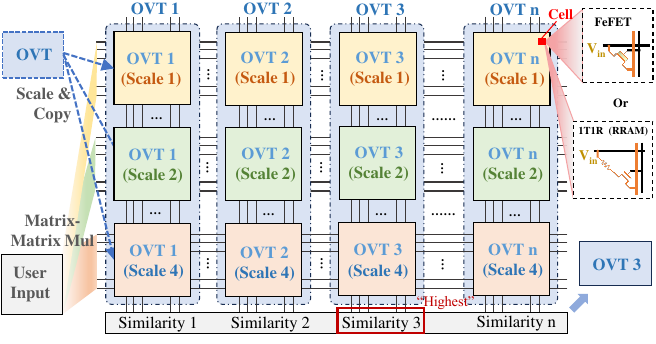}
  \captionsetup{font=small}  
  \caption{Implementation of our scaled search algorithm on NVCiM}
\vspace{-4.5ex}
  \label{fig:circuit}
\end{figure}



As shown in the lower-left block in Fig.~\ref{fig:method}, we employ a pre-trained autoencoder to transform the original embedding into an encoding space where the precision of each value in the embedding is compatible with NVM devices. The encoded embedding can be restored to its original precision via a decoder, allowing this restored embedding to be used in LLMs for content generation. 
Our autoencoder design follows the approach of \textit{Deep Compression}~\cite{han2015deep}.
Initially, we pre-train our autoencoder using user-generated data. The autoencoder is updated when the data buffer is full and representative data samples are selected for prompt tuning. The remaining data encircled with red dash-line in the blue block in Fig.~\ref{fig:method} is used to update the autoencoder. This updating process ensures that the autoencoder can properly encode and decode the user-generated data.

By default, our autoencoder encodes any OVTs into the shape with an embedding size of 48 and precision of \textit{int16}. We employ the subArray size of 384$\times$128  with 2-bit devices. As shown in Figure.~\ref{fig:circuit}, each square block can take one OVT, with the number of virtual tokens up to 128.

\newcolumntype{Y}{>{\centering\arraybackslash}X}
\newcolumntype{C}{>{\centering\arraybackslash}p{4.9mm}}  
\begin{table*}[htbp]
\fontsize{7pt}{5.5pt}\selectfont
\setlength{\tabcolsep}{1.8pt} 
\captionsetup{font=small}  
\caption{LLM average performance comparison between our framework and five baselines on 5 datasets, 3 edge LLMs, and 5 NVM devices, under the setting of 25 samples in each data buffer with device variation of 0.1. Settings are determined in TABLE~\ref{tab:baselines} and TABLE~\ref{tab:sigma}.}
\begin{tabularx}{\textwidth}{Cc|*{5}{Y}|*{5}{Y}|*{5}{Y}}
\toprule
 \multicolumn{2}{c}{\textbf{LLM}} & \multicolumn{5}{c}{\textbf{Gemma-2B}} & \multicolumn{5}{c}{\textbf{Mistral-7B-GPTQ}} & \multicolumn{5}{c}{\textbf{Phi-2}} \\
\midrule
\multicolumn{2}{c|}{\textbf{Dateset}} & \textbf{LaMP-1} & \textbf{LaMP-2} & \textbf{LaMP-3} & \textbf{LaMP-5} & \textbf{LaMP-7} & \textbf{LaMP-1} & \textbf{LaMP-2} & \textbf{LaMP-3} & \textbf{LaMP-5} & \textbf{LaMP-7} & \textbf{LaMP-1} & \textbf{LaMP-2} & \textbf{LaMP-3} & \textbf{LaMP-5} & \textbf{LaMP-7} \\
\textbf{Device} & \textbf{Method} & Acc & Acc & Acc & Rouge-1 & Rouge-1 & Acc & Acc & Acc & Rouge-1 & Rouge-1 & Acc & Acc & Acc & Rouge-1 & Rouge-1 \\
\midrule
\multirow{6}{*}{\rotatebox[origin=c]{90}{NVM-1}} 
& SWV       & 0.347 & 0.179 & 0.667 & 0.081 & 0.080 & 0.417 & 0.026 & 0.550 & 0.091 & 0.112 & 0.605 & 0.079 & 0.635 & 0.098 & 0.123  \\
& CxDNN     & 0.392 & 0.230 & 0.706 & 0.067 & 0.107 & 0.477 & 0.083 & 0.466 & 0.065 & 0.160 & 0.661 & 0.154 & 0.671 & 0.136 & 0.209  \\
& CorrectNet& 0.447 & 0.199 & 0.705 & 0.185 & 0.107 & 0.506 & 0.191 & 0.438 & 0.049 & 0.076 & 0.696 & 0.179 & 0.634 & 0.045 & 0.187  \\
& No-Miti(MIPS)& 0.318 & 0.153 & 0.634 & 0.128 & 0.081 & 0.421 & 0.011 & 0.466 & 0.03 & 0.011 & 0.593 & 0.148 & 0.627 & 0.052 & 0.148 \\
& NVP*(MIPS)& 0.385 & 0.155 & 0.668 & 0.020 & 0.051 & 0.393 & 0.111 & 0.251 & 0.053 & 0.118 & 0.409 & 0.104 & 0.354 & 0.118 & 0.216 \\
& NVCiM-PT & \textbf{0.549} & \textbf{0.250} & \textbf{0.732} & \textbf{0.199} & \textbf{0.139} & \textbf{0.529} & \textbf{0.205} & \textbf{0.559} & \textbf{0.166} & \textbf{0.209} & \textbf{0.707} & \textbf{0.250} & \textbf{0.725} & \textbf{0.201} & \textbf{0.257}  \\
\midrule
\multirow{6}{*}{\rotatebox[origin=c]{90}{NVM-2}} 
& SWV       & 0.360 & 0.119 & 0.702 & 0.107 & 0.044 & 0.341 & 0.038 & 0.554 & 0.105 & 0.160 & 0.691 & 0.083 & 0.681 & 0.085 & 0.049  \\
& CxDNN     & 0.359 & 0.305 & 0.755 & 0.151 & 0.124 & 0.487 & 0.063 & 0.452 & 0.117 & 0.101 & 0.667 & 0.227 & 0.687 & 0.085 & 0.255  \\
& CorrectNet& 0.469 & 0.209 & 0.685 & 0.226 & 0.147 & 0.440 & 0.107 & 0.507 & 0.147 & 0.093 & 0.672 & 0.207 & 0.643 & 0.138 & 0.128  \\
& No-Miti(MIPS)& 0.397 & 0.183 & 0.642 & 0.112 & 0.057 & 0.382 & 0.048 & 0.442 & 0.015 & 0.049 & 0.571 & 0.113 & 0.61 & 0.101 & 0.159  \\
& NVP*(MIPS)& 0.469 & 0.136 & 0.537 & 0.052 & 0.044 & 0.506 & 0.069 & 0.256 & 0.023 & 0.129 & 0.455 & 0.029 & 0.395 & 0.066 & 0.244  \\
& NVCiM-PT & \textbf{0.510} & \textbf{0.300} & \textbf{0.763} & \textbf{0.158} & \textbf{0.170} & \textbf{0.529} & \textbf{0.205} & \textbf{0.569} & \textbf{0.166} & \textbf{0.207} & \textbf{0.718} & \textbf{0.265} & \textbf{0.716} & \textbf{0.203} & \textbf{0.266}  \\
\midrule
\multirow{6}{*}{\rotatebox[origin=c]{90}{NVM-3}} 
& SWV       & 0.351 & 0.125 & 0.667 & 0.120 & 0.130 & 0.358 & 0.026 & 0.503 & 0.091 & 0.095 & 0.601 & 0.014 & 0.665 & 0.089 & 0.109 \\
& CxDNN     & 0.429 & 0.235 & 0.666 & 0.155 & 0.195 & 0.450 & 0.172 & 0.453 & 0.080 & 0.083 & 0.615 & 0.233 & 0.654 & 0.081 & 0.181  \\
& CorrectNet& 0.367 & 0.294 & 0.739 & 0.204 & 0.144 & 0.489 & 0.193 & 0.523 & 0.045 & 0.051 & 0.643 & 0.159 & 0.698 & 0.107 & 0.132  \\
& No-Miti(MIPS)& 0.386 & 0.233 & 0.704 & 0.082 & 0.066 & 0.419 & 0.026 & 0.474 & 0.035 & 0.048 & 0.520 & 0.056 & 0.612 & 0.054 & 0.099  \\
& NVP*(MIPS)& 0.470 & 0.119 & 0.548 & 0.123 & 0.109 & 0.447 & 0.167 & 0.238 & 0.060 & 0.145 & 0.473 & 0.072 & 0.362 & 0.001 & 0.265 \\
& NVCiM-PT & \textbf{0.500} & \textbf{0.320} & \textbf{0.765} & \textbf{0.225} & \textbf{0.198} & \textbf{0.529} & \textbf{0.205} & \textbf{0.559} & \textbf{0.167} & \textbf{0.204} & \textbf{0.696} & \textbf{0.265} & \textbf{0.725} & \textbf{0.185} & \textbf{0.267}  \\
\midrule
\multirow{6}{*}{\rotatebox[origin=c]{90}{NVM-4}}  
& SWV       & 0.353 & 0.163 & 0.697 & 0.050 & 0.055 & 0.347 & 0.042 & 0.458 & 0.045 & 0.187 & 0.673 & 0.008 & 0.607 & 0.068 & 0.040  \\
& CxDNN     & 0.408 & 0.270 & 0.721 & 0.113 & 0.209 & 0.500 & 0.187 & 0.541 & 0.097 & 0.049 & 0.610 & 0.157 & 0.614 & 0.071 & 0.214  \\
& CorrectNet& 0.466 & 0.216 & 0.701 & 0.183 & 0.107 & 0.430 & 0.096 & 0.578 & 0.061 & 0.113 & 0.691 & 0.140 & 0.662 & 0.171 & 0.231  \\
& No-Miti(MIPS)& 0.381 & 0.192 & 0.706 & 0.137 & 0.084 & 0.362 & 0.080 & 0.404 & 0.026 & 0.103 & 0.559 & 0.109 & 0.606 & 0.011 & 0.171  \\
& NVP*(MIPS)& 0.480 & 0.234 & 0.568 & 0.056 & 0.080 & 0.451 & 0.120 & 0.211 & 0.025 & 0.115 & 0.505 & 0.048 & 0.370 & 0.026 & 0.262  \\
& NVCiM-PT & \textbf{0.559} & \textbf{0.305} & \textbf{0.784} & \textbf{0.224} & \textbf{0.219} & \textbf{0.529} & \textbf{0.205} & \textbf{0.578} & \textbf{0.161} & \textbf{0.207} & \textbf{0.696} & \textbf{0.245} & \textbf{0.725} & \textbf{0.194} & \textbf{0.281}  \\
\midrule
\multirow{6}{*}{\rotatebox[origin=c]{90}{NVM-5}} 
& SWV       & 0.343 & 0.149 & 0.665 & 0.112 & 0.134 & 0.406 & 0.066 & 0.506 & 0.013 & 0.131 & 0.625 & 0.094 & 0.632 & 0.081 & 0.092  \\
& CxDNN     & 0.492 & 0.255 & 0.701 & 0.098 & 0.105 & 0.389 & 0.153 & 0.529 & 0.042 & 0.069 & 0.660 & 0.143 & 0.640 & 0.061 & 0.247  \\
& CorrectNet& 0.435 & 0.299 & 0.724 & 0.203 & 0.171 & 0.463 & 0.175 & 0.477 & 0.014 & 0.162 & 0.619 & 0.138 & 0.694 & 0.095 & 0.268  \\
& No-Miti(MIPS)& 0.420 & 0.157 & 0.702 & 0.102 & 0.068 & 0.396 & 0.021 & 0.389 & 0.036 & 0.052 & 0.517 & 0.151 & 0.660 & 0.058 & 0.155  \\
& NVP*(MIPS)& 0.412 & 0.189 & 0.563 & 0.109 & 0.131 & 0.423 & 0.167 & 0.196 & 0.015 & 0.139 & 0.447 & 0.012 & 0.463 & 0.117 & 0.204  \\
& NVCiM-PT & \textbf{0.608} & \textbf{0.310} & \textbf{0.775} & \textbf{0.252} & \textbf{0.216} & \textbf{0.529} & \textbf{0.205} & \textbf{0.569} & \textbf{0.168} & \textbf{0.203} & \textbf{0.755} & \textbf{0.265} & \textbf{0.745} & \textbf{0.202} & \textbf{0.270}  \\
\bottomrule
\end{tabularx}
\vspace{-3ex}
\label{tab:main}
\end{table*}

\subsubsection{Retrieval of OVTs}

After the OVTs are encoded and stored on NVM devices, the next step is to retrieve them efficiently based on the user input. To implement the retrieval, all stored OVTs need to be compared with the user input. A possible way is to apply the MIPS between the embedding of the user input and the stored OVTs \cite{vor2019text}.
However, the format of user inputs represented purely as word embedding \cite{mars2022word}, is not compatible with that of virtual tokens obtained from prompt tuning, since the virtual tokens can contain additional information like task domain information \cite{xu2023parameter}. While this incompatibility issue can reduce the effectiveness of MIPS as shown in our experiments in Section~\ref{sec:exp}, this issue can be mitigated by pooling \cite{shen2018baseline}. Hence, we employ a multi-level scaling strategy based on average pooling into our search algorithm, which we refer to SSA.


Our method leverages the user input and stored OVTs by employing average pooling based on three scale factors: 1, 2, and 4. This mitigates the incompatibility between the two types of matrices. As shown in Fig.~\ref{fig:circuit}, the yellow, green, and orange colors represent these scale factors respectively. For each set of virtual tokens $p$ in the set of all prompts $P$, we store its three scaled versions. The algorithm, based on General Matrix Multiplication (GMM) employing the CiM architecture, finds the most similar sets of virtual tokens $p^*$ to the user inputs $e$ using a Weighted Multi-Scale Dot Product (WMSDP) function, where \(p^* = \underset{p \in P}{\text{argmax}} \text{ WMSDP}(e, p)\) and WMSDP is defined as:

{\footnotesize
\begin{equation}
\text{WMSDP}(e, p) = \frac{\sum_{i \in L} w_i \cdot (\text{Pool}_i(e) \cdot \text{Pool}_i(p))}{\sum{i \in L} w_i}
\end{equation}
}

\noindent The argmax operation involves computing the weighted sum of matrix multiplications at different scales. \rqin{The WMSDP produces a similarity score between the user input and each stored OVT as shown in Fig.~\ref{fig:circuit}.} For scales, $L = {1, 2, 4}$ is the set of scales with corresponding weights ${W_1, W_2, W_3} = {1, 0.8, 0.6}$. The dot product $(a \cdot b)$ is used for similarity, and the pooling operation $\text{Pool}_i(x) = \frac{1}{i} \sum{j=1}^{i} x{k:k+i}$ for $k = 1, 1+i, 1+2i, \ldots$, where $x$ is either $e$ or $p$. The scaled search captures similarities at different granularities: token-level (scale 1), adjacent token pairs (scale 2), and broader contextual level (scale 4). The weighted average ensures that fine-grained similarities have a higher impact than the coarser-grained ones while all scales contribute to the similarity score.

\section{Experimental Evaluation}
\label{sec:exp}
\subsection{Experimental Setup}

\newcolumntype{Y}{>{\centering\arraybackslash}X}

\begin{table}[t]
\fontsize{7pt}{5pt}\selectfont
    \centering
    \captionsetup{font=small}  
    \caption{Device non-ideality modeling for different real and synthesized devices. For devices with more than two levels, the device variation for each level is depicted as $L_x$.}
    \begin{tabularx}{\columnwidth}{c*{5}{Y}}        
    \toprule
        \multirow{2}{*}{Name} & \multirow{2}{*}{\makecell{\# of \\ Levels}}  & \multicolumn{4}{c}{Device Variations $\sigma_v$} \\
                  &   & $L_0$ & $L_1$ & $L_2$ & $L_3$ \\
        \midrule
        $RRAM_1$ (NVM-1)  & 1 & 0.0100 & 0.0100 & 0.0100 & 0.0100\\
        $FeFET_2$ (NVM-2) & 4 & 0.0067 & 0.0135 & 0.0135 & 0.0067\\
        $FeFET_3$ (NVM-3) & 4 & 0.0049 & 0.0146 & 0.0146 & 0.0049\\
        $RRAM_4$ (NVM-4)  & 4 & 0.0038 & 0.0151 & 0.0151 & 0.0038\\
        $FeFET_6$ (NVM-5) & 4 & 0.0026 & 0.0155 & 0.0155 & 0.0026\\
        \bottomrule
    \end{tabularx}
    \vspace{-5ex}
    \label{tab:var}
\end{table}

\subsubsection{Datasets}
We select three classification datasets including LaMP-1, LaMP-2, and LaMP-3, and two summarization datasets including LaMP-5 and LaMP-7 from LaMP datasets to evaluate our \textbf{NVCiM-PT} framework. Each LaMP dataset has over 100 users and contains over 100 data samples. We use the average performance of over 100 users in each LaMP dataset as the final experimental results. 

\subsubsection{Default Experimental Setting}
\label{sec:default_setting}

In our experiments, we first select three LLMs including Gemma-2B, Mistral-7B-GPTQ, and Phi2, 
 where these models can be deployed on edge devices.
For each model, we set the temperature to be \(0.1\) so we have high certainty during content generation. We also limit the number of generated tokens to 100 to avoid long and trivial LLM-generated content. For obtaining the OVT of each data sample under each selected LLM, we use the HuggingFace hosted original PT \cite{lester2021power}. We set its learning rate to be \(1e-4\), corresponding to the \textit{adam} optimizer with a learning rate scheduler. For NVM device variation, we set its standard deviation \(\sigma\) to be 0.1. In the following experiments, if no specific explanations, it indicates that we use the default settings.


In all experiments, we adhere to the device variation model previously described. The specific parameters are abstracted and then simplified from three representative NVM devices, two of them are resistive random-access memory (RRAM) devices extracted from~\cite{yao2020fully, liu2023architecture}, and the other is a ferroelectric field effect transistor (FeFET) device extracted from~\cite{wei2022switching}. We name them $RRAM_1$, $RRAM_4$ and $FeFET_2$, respectively. We also extrapolate the modeling data to obtain two synthesized $FeFET_3$ and $FeFET_6$ devices. Detailed device modeling results are demonstrated in Table~\ref{tab:var}. An $x$-level device means that this device can represent $x$ distinct values and $\sigma_{L_2} = 0.01$ means the variation of this device is 0.01 when it is representing the 2nd level value 1. These selected devices represent a diverse range of NVM technologies and performance characteristics \cite{yan2022swim, yan2021uncertainty}, providing a comprehensive basis for evaluating CiM architectures across different device scenarios. 


\subsubsection{Evaluation Methods}

We employed two performance evaluation metrics: Accuracy and ROUGE-1. Accuracy assesses the model's ability to correctly classify instances across different categories, which is particularly relevant for the classification tasks in LaMP-1, LaMP-2, and LaMP-3. ROUGE-1 \cite{lin2004rouge} evaluates the word overlap between generated and reference texts, crucial for the text generation tasks in LaMP-5 and LaMP-7. Using them, we comprehensively assessed our framework's performance across various task types, including binary classification, multiclass classification, and text generation.

\subsubsection{Baselines}

In Fig.~\ref{tab:one4all}, we have already compared the performance of prompt tuning relying on the one4all set of virtual tokens and the prompt tuning method when the OVT can be trained for every data sample. Hence, we concentrate the rest of the experiments on NVM devices. For the noise mitigation methods, we select three SOTA works including selective write verify (SWV) \cite{yan2022swim}, CxDNN \cite{jain2019cxdnn}, and CorrectNet \cite{eldebiky2023correctnet}. For the OVTs processed by them and stored in NVMs, we still use our proposed SSA to retrieve OVTs. To evaluate the performance of our SSA, we compare it with MIPS \cite{shen2015learning} in two ways, i.e., no noise mitigation (No-Miti) and NVCiM-PT with MIPS (NVP*(MIPS)) instead of our SSA.

\begin{table}[t!]
\captionsetup{font=small}  
\caption{Performance comparison between our framework and baselines on NVCiM-3 with device variation \(\sigma = 0.1\), using Phi-2 on LaMP-5, with different data buffer sizes }
\centering
\fontsize{7pt}{6pt}\selectfont
\setlength{\tabcolsep}{3pt}  
\begin{tabularx}{\columnwidth}{>{\centering\arraybackslash}m{1.5cm}|*{6}{>{\centering\arraybackslash}m{1cm}}}
\toprule
Buffer Size & SWV & CxDNN & CorrectNet & No-Miti (MIPS) & NVP* (MIPS) & NVCiM-PT \\ 
\midrule
10 samples & 0.153 & 0.153 & 0.151 & 0.151 & 0.153 & \textbf{0.158} \\
20 samples & 0.149 & 0.149 & 0.159 & 0.141 & 0.166 & \textbf{0.207} \\
30 samples & 0.137 & 0.137 & 0.177 & 0.134 & 0.181 & \textbf{0.241} \\
40 samples & 0.171 & 0.171 & 0.177 & 0.132 & 0.171 & \textbf{0.182} \\
50 samples & 0.156 & 0.156 & 0.171 & 0.130 & 0.180 & \textbf{0.199} \\
60 samples & 0.144 & 0.144 & 0.172 & 0.135 & 0.179 & \textbf{0.222} \\
\bottomrule
\end{tabularx}
\vspace{-3ex}
\label{tab:baselines}
\end{table}

\subsection{Results}
\subsubsection{Performance Studies}
In our experiments, we first evaluate the impact of representative selection on our NVCiM-PT framework. For this study, we vary the buffer size and evaluate how our framework performs across different LLMs and different NVM devices. Since measuring whether the retrieved NVM-stored OVT is optimal corresponding to the user input can be hard (i.e., calculate accuracy based on labels that do not exist in our setting), we directly evaluate the LLM-generated content accuracy or ROUGE-1, where higher values stand for better NVM-stored OVT retrieval.

\begin{table}[htbp]  
\captionsetup{font=small}  
\caption{Performance comparison between our framework and baselines on NVCiM-3 with different device variations \(\sigma\), using LLM Phi-2 on LaMP-5, given data buffer size 20}
\centering
\fontsize{7pt}{6pt}\selectfont
\setlength{\tabcolsep}{3pt}  
\begin{tabularx}{\columnwidth}{>{\centering\arraybackslash}m{1.5cm}|*{6}{>{\centering\arraybackslash}m{1cm}}}
\toprule
Dev. var. ($\sigma$) & SWV & CxDNN & CorrectNet & No-Miti (MIPS) & NVP* (MIPS) & NVCiM-PT \\ 
\midrule
0.025 & 0.149 & 0.150 & 0.159 & 0.141 & 0.166 & \textbf{0.215}  \\
0.050 & 0.147 & 0.145 & 0.158 & 0.139 & 0.167 & \textbf{0.206}  \\
0.075 & 0.148 & 0.140 & 0.160 & 0.140 & 0.165 & \textbf{0.191}  \\
0.100 & 0.149 & 0.149 & 0.159 & 0.141 & 0.166 & \textbf{0.207}  \\
0.125 & 0.146 & 0.146 & 0.156 & 0.137 & 0.163 & \textbf{0.185}  \\
0.150 & 0.143 & 0.141 & 0.155 & 0.135 & 0.162 & \textbf{0.189}  \\
\bottomrule
\end{tabularx}
\vspace{-3ex}
\label{tab:sigma}
\end{table}

As shown in TABLE~\ref{tab:baselines}, we initialize the data buffer size from 10 to 70. Under each setting, we compare the three noise mitigation methods and two cases using MIPS, with our NVCiM-PT framework. In this experiment, we use the dataset of LaMP-5 and choose Phi-2 as the edge LLM. The results in this table demonstrate that our NVCiM-PT in general has the best performance across all buffer sizes. Notably, its performance under the size of 30 is the best, indicating that it can be beneficial to choose a medium buffer size. In addition, the performance of NVM*(MIPS) also outperforms the three noise mitigation methods and ``No-Miti(MIPS)". Together it can be seen that our noise-aware training, in conjunction with our SSA provides the best performance. 

Furthermore, we examine the impact of device variations ranging from \(\sigma=0.025\) to \(\sigma=0.150\), as shown in TABLE~\ref{tab:sigma}. Our experiments demonstrate that our framework consistently outperforms baselines across all device variations, despite a slight performance degradation observed in both our framework and the baselines as variation increases.

After we study the impact of buffer size (TABLE~\ref{tab:baselines}), and then the effect of noise-aware training for variations with different standard variations (TABLE~\ref{tab:sigma}), we further conduct a comprehensive evaluation between our NVCiM-PT and the baselines. As we mentioned in the Section~\ref{sec:NVM}, we select five NVM devices and use their device variation settings as shown in TABE~\ref{tab:var} with \(\sigma = 0.1\) to examine whether NVCiM-PT can work properly with different NVM devices. As shown in TABLE~\ref{tab:main}, under each LLM, we compare five baselines with our NVCiM-PT on five NVM devices across five datasets. Among the five NVM devices and three LLMs, NVCiM-PT outperforms these baselines. Specifically, NVP*(MIPS) demonstrates the usefulness of our SSA and No-Miti(MIPS) demonstrates the usefulness of our noise-aware training method. Furthermore, the three noise mitigation methods (SWV, CxDNN, and CorrectNet) together with our SSA can outperform No-Miti(MIPS) demonstrating that our SSA can help with a wide range of noise mitigation methods when they are used to store data into NVMs. 

\begin{figure}[t!]
  \centering
  \begin{subfigure}[b]{0.24\textwidth}
    \centering
    \includegraphics[width=\textwidth]{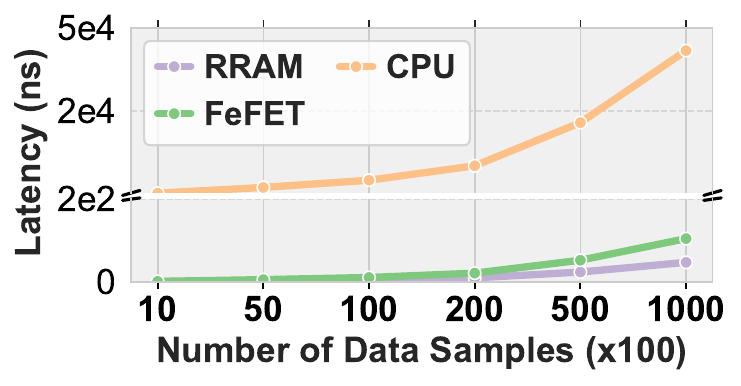}
    \vspace{-4ex}
    \caption{Search Latency}
    
    \label{fig:latency}
  \end{subfigure}
  \begin{subfigure}[b]{0.24\textwidth}
    \centering
    \includegraphics[width=\textwidth]{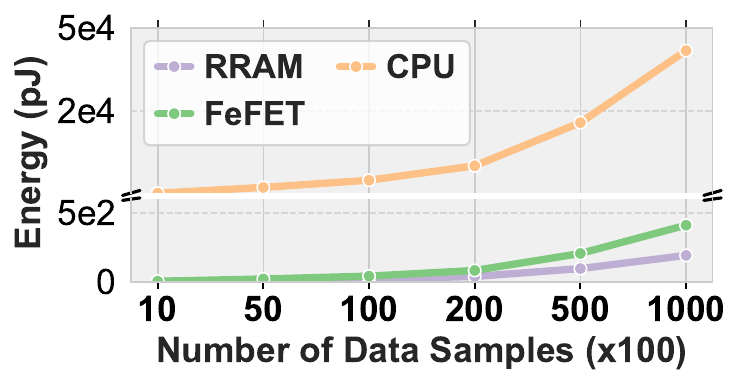}
    \vspace{-4ex}
    \caption{Energy cost}
    
    \label{fig:energy}
  \end{subfigure}
  \captionsetup{font=small}  
  \caption{Evaluation of our SSA on CPU, RRAM, and FeFET}
  \vspace{-4ex}
  \label{fig:ablation_2}
\end{figure}

\subsubsection{Latency \& energy studies}
We also evaluate the latency and energy of retrieval of the appropriate data by our scaled search algorithm on NVCiM via NeuroSim \cite{peng2020dnn+}. In our simulation, \textcolor{black}{we obtain the latency and energy of crossbar array and peripheral circuits for the 22nm technology node.} We evaluate our frameowrk on two NVCiM devices including RRAM and FeFET, and Jetson Orin CPU.
As shown in Fig.~\ref{fig:ablation_2}, our algorithm on NVCiM devices can have significantly lower latency and energy, validating the benefit of NVCiM in our framework.

\section{Conclusion}
In this paper, we present a novel framework NVCiM-PT, which exploits NVCiM to assist prompt tuning for edge LLMs. By properly selecting the representative data samples, training the OVTs from them, and retrieving the appropriate OVT in NVCiM, our approach provides a solution to efficiently store and retrieve the OVTs for a small range of user-generated data. Experimental results show that NVCiM-PT outperforms existing noise mitigation works and the commonly used MIPS search method on various NVM devices.

\clearpage
\bibliographystyle{unsrt}
\bibliography{citations}

\begin{thebibliography}{10}

\bibitem{qin2024fl}
Ruiyang Qin, Yuting Hu, Zheyu Yan, Jinjun Xiong, Ahmed Abbasi, and Yiyu Shi.
\newblock Fl-nas: Towards fairness of nas for resource constrained devices via large language models.
\newblock {\em arXiv preprint arXiv:2402.06696}, 2024.

\bibitem{qin2024language}
Ruiyang Qin, Ryan Cook, Kai Yang, Ahmed Abbasi, David Dobolyi, Salman Seyedi, Emily Griner, Hyeokhyen Kwon, Robert Cotes, Zifan Jiang, et~al.
\newblock Language models for online depression detection: A review and benchmark analysis on remote interviews.
\newblock {\em ACM Transactions on Management Information Systems}, 2024.

\bibitem{bevilacqua2023automated}
Marialena Bevilacqua, Kezia Oketch, Ruiyang Qin, Will Stamey, Xinyuan Zhang, Yi~Gan, Kai Yang, and Ahmed Abbasi.
\newblock When automated assessment meets automated content generation: Examining text quality in the era of gpts.
\newblock {\em arXiv preprint arXiv:2309.14488}, 2023.

\bibitem{neel2023privacy}
Seth Neel and Peter Chang.
\newblock Privacy issues in large language models: A survey, 2023.

\bibitem{Karabacak_Margetis_2023}
Karabacak et~al.
\newblock Embracing large language models for medical applications: Opportunities and challenges.
\newblock {\em Cureus}, May 2023.

\bibitem{xu2023large}
Xu~et~al.
\newblock Can large language models be good companions? an llm-based eyewear system with conversational common ground, 2023.

\bibitem{li2024personal}
Li~et~al.
\newblock Personal llm agents: Insights and survey about the capability, efficiency and security, 2024.

\bibitem{qin2024robust}
Ruiyang Qin, Zheyu Yan, Dewen Zeng, Zhenge Jia, Dancheng Liu, Jianbo Liu, Zhi Zheng, Ningyuan Cao, Kai Ni, Jinjun Xiong, et~al.
\newblock Robust implementation of retrieval-augmented generation on edge-based computing-in-memory architectures.
\newblock {\em arXiv preprint arXiv:2405.04700}, 2024.

\bibitem{qin2023enabling}
Ruiyang Qin, Jun Xia, Zhenge Jia, Meng Jiang, Ahmed Abbasi, Peipei Zhou, Jingtong Hu, and Yiyu Shi.
\newblock Enabling on-device large language model personalization with self-supervised data selection and synthesis.
\newblock {\em arXiv preprint arXiv:2311.12275}, 2023.

\bibitem{qin2024empirical}
Ruiyang Qin, Dancheng Liu, Zheyu Yan, Zhaoxuan Tan, Zixuan Pan, Zhenge Jia, Meng Jiang, Ahmed Abbasi, Jinjun Xiong, and Yiyu Shi.
\newblock Empirical guidelines for deploying llms onto resource-constrained edge devices.
\newblock {\em arXiv preprint arXiv:2406.03777}, 2024.

\bibitem{li2021prefix}
Xiang~Lisa Li and Percy Liang.
\newblock Prefix-tuning: Optimizing continuous prompts for generation.
\newblock {\em arXiv preprint arXiv:2101.00190}, 2021.

\bibitem{shi2023dept}
Zhengxiang Shi and Aldo Lipani.
\newblock Dept: Decomposed prompt tuning for parameter-efficient fine-tuning.
\newblock {\em arXiv preprint arXiv:2309.05173}, 2023.

\bibitem{nassereldine2024pi}
Amir Nassereldine, Dancheng Liu, Chenhui Xu, and Jinjun Xiong.
\newblock Pi-whisper: An adaptive and incremental asr framework for diverse and evolving speaker characteristics.
\newblock {\em arXiv preprint arXiv:2406.15668}, 2024.

\bibitem{lester2021power}
Brian Lester, Rami Al-Rfou, and Noah Constant.
\newblock The power of scale for parameter-efficient prompt tuning.
\newblock {\em arXiv preprint arXiv:2104.08691}, 2021.

\bibitem{liu2021p}
Xiao Liu, Kaixuan Ji, Yicheng Fu, Weng~Lam Tam, Zhengxiao Du, Zhilin Yang, and Jie Tang.
\newblock P-tuning v2: Prompt tuning can be comparable to fine-tuning universally across scales and tasks.
\newblock {\em arXiv preprint arXiv:2110.07602}, 2021.

\bibitem{yin2024ferroelectric}
Xunzhao Yin, Yu~Qian, Alptekin Vardar, Marcel G{\"u}nther, Franz M{\"u}ller, Nellie Laleni, Zijian Zhao, Zhouhang Jiang, Zhiguo Shi, Yiyu Shi, et~al.
\newblock Ferroelectric compute-in-memory annealer for combinatorial optimization problems.
\newblock {\em Nature Communications}, 15(1):2419, 2024.

\bibitem{jung2022crossbar}
Seungchul Jung, Hyungwoo Lee, Sungmeen Myung, Hyunsoo Kim, Seung~Keun Yoon, Soon-Wan Kwon, Yongmin Ju, Minje Kim, Wooseok Yi, Shinhee Han, et~al.
\newblock A crossbar array of magnetoresistive memory devices for in-memory computing.
\newblock {\em Nature}, 601(7892):211--216, 2022.

\bibitem{wan2022compute}
Weier Wan, Rajkumar Kubendran, Clemens Schaefer, Sukru~Burc Eryilmaz, Wenqiang Zhang, Dabin Wu, Stephen Deiss, Priyanka Raina, He~Qian, Bin Gao, et~al.
\newblock A compute-in-memory chip based on resistive random-access memory.
\newblock {\em Nature}, 608(7923):504--512, 2022.

\bibitem{khan2020future}
Asif~Islam Khan, Ali Keshavarzi, and Suman Datta.
\newblock The future of ferroelectric field-effect transistor technology.
\newblock {\em Nature Electronics}, 3(10):588--597, 2020.

\bibitem{yan2022swim}
Yan et~al.
\newblock Swim: Selective write-verify for computing-in-memory neural accelerators.
\newblock In {\em 2022 59th ACM/IEEE Design Automation Conference (DAC)}. IEEE.

\bibitem{yan2021uncertainty}
Yan et~al.
\newblock Uncertainty modeling of emerging device based computing-in-memory neural accelerators with application to neural architecture search.
\newblock In {\em 2021 26th Asia and South Pacific Design Automation Conference (ASP-DAC)}. IEEE, 2021.

\bibitem{gong2018signal}
Nanbo Gong, T~Id{\'e}, S~Kim, Irem Boybat, Abu Sebastian, Vijay Narayanan, and Takashi Ando.
\newblock Signal and noise extraction from analog memory elements for neuromorphic computing.
\newblock {\em Nature communications}, 9(1):2102, 2018.

\bibitem{yan2022reliability}
Zheyu Yan, Xiaobo~Sharon Hu, and Yiyu Shi.
\newblock On the reliability of computing-in-memory accelerators for deep neural networks.
\newblock In {\em System Dependability and Analytics: Approaching System Dependability from Data, System and Analytics Perspectives}, pages 167--190. Springer, 2022.

\bibitem{han2015deep}
Song Han, Huizi Mao, and William~J Dally.
\newblock Deep compression: Compressing deep neural networks with pruning, trained quantization and huffman coding.
\newblock {\em arXiv preprint arXiv:1510.00149}, 2015.

\bibitem{vor2019text}
Tim vor~der Br{\"u}ck and Marc Pouly.
\newblock Text similarity estimation based on word embeddings and matrix norms for targeted marketing.
\newblock In {\em Proceedings of the 2019 Conference of the North American Chapter of the Association for Computational Linguistics: Human Language Technologies, Volume 1 (Long and Short Papers)}, pages 1827--1836, 2019.

\bibitem{mars2022word}
Mourad Mars.
\newblock From word embeddings to pre-trained language models: A state-of-the-art walkthrough.
\newblock {\em Applied Sciences}, 12(17):8805, 2022.

\bibitem{xu2023parameter}
Lingling Xu, Haoran Xie, Si-Zhao~Joe Qin, Xiaohui Tao, and Fu~Lee Wang.
\newblock Parameter-efficient fine-tuning methods for pretrained language models: A critical review and assessment.
\newblock {\em arXiv preprint arXiv:2312.12148}, 2023.

\bibitem{shen2018baseline}
Dinghan Shen, Guoyin Wang, Wenlin Wang, Martin~Renqiang Min, Qinliang Su, Yizhe Zhang, Chunyuan Li, Ricardo Henao, and Lawrence Carin.
\newblock Baseline needs more love: On simple word-embedding-based models and associated pooling mechanisms.
\newblock {\em arXiv preprint arXiv:1805.09843}, 2018.

\bibitem{yao2020fully}
Yao et~al.
\newblock Fully hardware-implemented memristor convolutional neural network.
\newblock {\em Nature}, 577(7792):641--646, 2020.

\bibitem{liu2023architecture}
Liu et~al.
\newblock Architecture-circuit-technology co-optimization for resistive random access memory-based computation-in-memory chips.
\newblock {\em Science China Information Sciences}, 66(10):200408, 2023.

\bibitem{wei2022switching}
Wei et~al.
\newblock Switching pathway-dependent strain-effects on the ferroelectric properties and structural deformations in orthorhombic hfo2.
\newblock {\em Journal of Applied Physics}, 131(15), 2022.

\bibitem{lin2004rouge}
Chin-Yew Lin.
\newblock Rouge: A package for automatic evaluation of summaries.
\newblock In {\em Text summarization branches out}, pages 74--81, 2004.

\bibitem{jain2019cxdnn}
Shubham Jain and Anand Raghunathan.
\newblock Cxdnn: Hardware-software compensation methods for deep neural networks on resistive crossbar systems.
\newblock {\em ACM Transactions on Embedded Computing Systems (TECS)}, 18(6):1--23, 2019.

\bibitem{eldebiky2023correctnet}
Amro Eldebiky, Grace~Li Zhang, Georg B{\"o}cherer, Bing Li, and Ulf Schlichtmann.
\newblock Correctnet: Robustness enhancement of analog in-memory computing for neural networks by error suppression and compensation.
\newblock In {\em 2023 Design, Automation \& Test in Europe Conference \& Exhibition (DATE)}, pages 1--6. IEEE, 2023.

\bibitem{shen2015learning}
Fumin Shen, Wei Liu, Shaoting Zhang, Yang Yang, and Heng Tao~Shen.
\newblock Learning binary codes for maximum inner product search.
\newblock In {\em Proceedings of the IEEE International Conference on Computer Vision}, pages 4148--4156, 2015.

\bibitem{peng2020dnn+}
Xiaochen Peng, Shanshi Huang, Hongwu Jiang, Anni Lu, and Shimeng Yu.
\newblock Dnn+ neurosim v2. 0: An end-to-end benchmarking framework for compute-in-memory accelerators for on-chip training.
\newblock {\em IEEE Transactions on Computer-Aided Design of Integrated Circuits and Systems}, 40(11):2306--2319, 2020.

\end{thebibliography}

\end{document}